\documentclass[accepted, table, xcdraw]{article}

\usepackage{microtype, pdfpages, enumerate}
\usepackage{graphicx}
\usepackage{subfigure}
\usepackage{booktabs} 

\usepackage{hyperref}
\usepackage{array}
\newcolumntype{C}[1]{>{\arraybackslash}p{#1}}

\usepackage[accepted]{icml2024}

\usepackage{amsmath}
\usepackage{amssymb}
\usepackage{mathtools}
\usepackage{amsthm, multirow}
\usepackage{xcolor}
\usepackage[normalem]{ulem}
\useunder{\uline}{\ul}{}

\usepackage[capitalize,noabbrev]{cleveref}

\theoremstyle{plain}

\theoremstyle{definition}

\theoremstyle{remark}

\usepackage[textsize=tiny]{todonotes}

\icmltitlerunning{RAG Evaluation Telecom QA}

\begin{document}

\twocolumn[
\icmltitle{Evaluation of RAG Metrics for Question Answering in the Telecom Domain}



\icmlsetsymbol{equal}{*}

\begin{icmlauthorlist}
\icmlauthor{Sujoy Roychowdhury}{equal,yyy}
\icmlauthor{Sumit Soman}{equal,yyy}
\icmlauthor{H G Ranjani}{equal,yyy}
\icmlauthor{Neeraj Gunda}{sch}
\icmlauthor{Vansh Chhabra}{sch}
\icmlauthor{Sai Krishna Bala}{yyy}
\end{icmlauthorlist}

\icmlaffiliation{yyy}{Ericsson R\&D, Bangalore, India}
\icmlaffiliation{sch}{Work done during internship at Ericsson R\&D}
\icmlcorrespondingauthor{Sujoy Roychowdhury, Sumit Soman, Ranjani HG}{sujoy.roychowdhury, sumit.soman, ranjani.h.g@ericsson.com}

\icmlkeywords{Machine Learning, ICML}

\vskip 0.3in
]



\printAffiliationsAndNotice{Code is available at \url{https://anonymous.4open.science/r/ragas_updated-FFC6}}. 

\begin{abstract}
Retrieval Augmented Generation (RAG) is widely used to enable Large Language Models (LLMs) perform Question Answering (QA) tasks in various domains. However, RAG based on open-source LLM for specialized domains has challenges of evaluating generated responses. A popular framework in the literature is the RAG Assessment (RAGAS), a publicly available library which uses LLMs for evaluation. One disadvantage of RAGAS is the lack of details of derivation of numerical value of the evaluation metrics. One of the outcomes of this work is a modified version of this package for few metrics (faithfulness, context relevance, answer relevance, answer correctness, answer similarity and factual correctness) through which we provide the intermediate outputs of the prompts by using any LLMs. Next, we analyse the expert evaluations of the output of the modified RAGAS package and observe the challenges of using it in the telecom domain. We also study the effect of the metrics under correct vs. wrong retrieval and observe that few of the metrics have higher values for correct retrieval. We also study for differences in metrics between base embeddings and those domain adapted via pre-training and fine-tuning. Finally, we comment on the suitability and challenges of using these metrics for in-the-wild telecom QA task.
\end{abstract}


\section{Introduction}
Retrieval Augmented Generation (RAG) \cite{lewis2020retrieval} is one of the approaches to enable Question Answering (QA) from specific domains, while leveraging generative capabilities of Large Language Models (LLMs). There have been many techniques proposed to enhance RAG performance such as chunk length, order of retrieved chunks in context \cite{chen2023understanding,soman2024observations}. 
However, like all systems, these require objective metrics to measure performance of the end-to-end system. The challenge in evaluating RAG system lies in comparing the generated answer with the ground truth for factualness, relevance to question and semantic similarity \cite{chen2024benchmarking}.

Initial approaches for RAG evaluation included re-purposing metrics used for machine translation tasks such as BLEU \cite{papineni2002bleu}, ROUGE \cite{lin2004rouge} or METEOR \cite{banerjee2005meteor}. Text generation was also evaluated using BERTScore \cite{zhang2019bertscore}. Metrics from Natural Language Inference (MeNLI) \cite{chen2023menli} built an adversarial attack framework to demonstrate improvements over BERTScore. Classical methods used for evaluation such as Item Response Theory have also been explored for RAG evaluation \cite{guinet2024automated}. The limitations with these techniques are: (i) they have a limited contextual input, (ii) can potentially look for either exact matches or semantic similarity aspects only, and (iii) are measured at sentence level only. RAG response evaluations, however, require a combination of exact match for factual component(s) and semantic similarities for relevance. 

In an attempt to mimic human intuition in assessing logical and grounded conversations, metrics based on prompting LLMs to evaluate RAG outputs have been proposed. RAG Assessment (RAGAS) \cite{es2023ragas} proposes multiple measures such as faithfulness, context and answer relevance to assess RAG responses using specific prompts.  We use this framework for assessment as it is one of the first and has also been used in popular courses \cite{LiuDatta2024}.


Our work is motivated by the need to evaluate these metrics for RAG systems in technical domains; we focus on telecom as an example. Most of the prior art focuses on evaluation using public datasets \cite{yang2024crag}. However, with increasing applications that use LLMs for telecom \cite{zhou2024large, karapantelakis2024using, soman2023observations}, it is important to assess the robustness of these metrics in the presence of domain-specific terminology. Further, there can be potential improvements in the RAG pipeline, such as domain adaptation of the retriever or instruction tuning of the LLM. Our work examines the effect of some of these on RAG metrics, in order to assess their adequacy and effectiveness. 
Our study serves as a starting point for evaluation of baseline RAGAS metrics for technical QA in telecom domain. Specifically, in this study, we consider the following metrics: (i) Faithfulness (ii) Answer Relevance (iii) Context Relevance (iv) Answer Similarity (v) Factual Correctness, and (vi) Answer Correctness.

\subsection{Research Questions}
The Research Questions (RQs) considered in this work are:
\begin{itemize}
    \item \textbf{RQ1:} How do LLM-based evaluation metrics, specifically RAGAS, go through the step-by-step evaluation procedure specified by the prompts? 
    \item \textbf{RQ2:} Are RAGAS metrics appropriate for evaluation of telecom QA task using RAG?
    \item \textbf{RQ3:} Are RAGAS metrics affected by retriever performance, domain adapted embeddings and instruction tuned LLM.
\end{itemize}

The contributions of our work are as follows:
\begin{enumerate}
    \item We have enhanced the RAGAS public repository code by capturing the intermediate outputs of all the prompts used to compute RAGAS metrics. This provides better visibility of inner workings of the metrics and possibility to modify the prompts.
    \item We manually evaluate, for \textbf{RQ1}, the intermediate outputs of the considered metrics with respect to the context and ground truth. We critically analyse them for their appropriateness for RAG using telecom domain data. 
    \item We establish, for \textbf{RQ2}, that two of the metrics - Factual Correctness and Faithfulness, are good indicators of correctness of the RAG response with respect to expert evaluation. We demonstrate that use of these two metrics together is better at identifying correctness of response; this improves further on using domain adapted LLMs.
    \item We establish, for \textbf{RQ3}, that Factual Correctness metric improves with instruction fine tuning of generator LLMs, irrespective of retrieved context. We observe lower Faithfulness metric for RAG answers which are identified to be correct but from wrong retrieved context. This indicates that the generator (LLM) has answered from out of context information. The ability to answer from out-of-context information is more pronounced for domain adapted generator. Thus, the metrics are able to reflect the expected negative correlation between faithfulness and factual correctness for wrong retrieval - although ideally the RAG system should have not provided an answer for a wrong retrieved context.

\end{enumerate}


\section{Experimental Setup}
\label{sec:experimental_setup}
\subsection{Dataset}
\begin{figure}[t!]
    \centering
\includegraphics[width=0.98\linewidth]{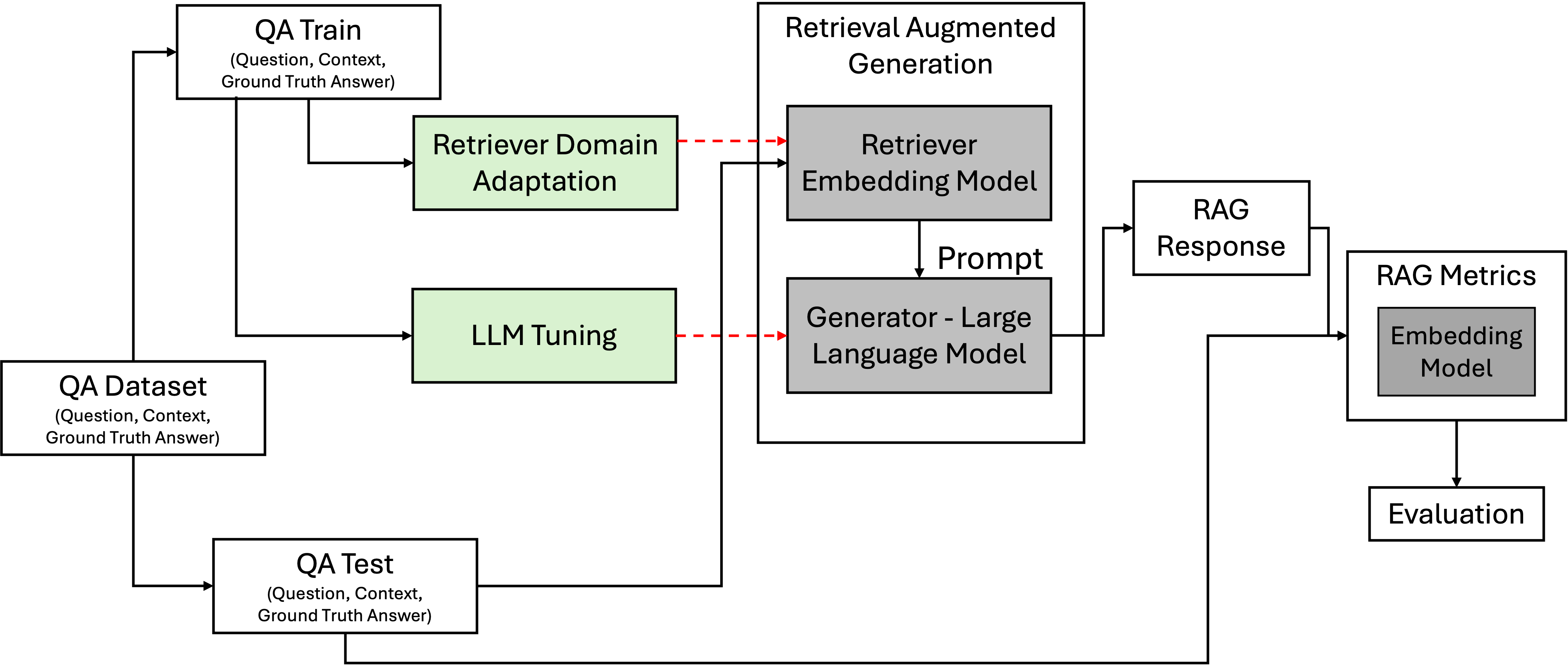}
    \caption{Schematic showing our experimental setup. Dotted arrows indicate that the retriever and generator are evaluated with both the base and domain adapted variants.}
    \label{fig:expt-setup}
\end{figure}

All experiments in this work are based on subset of TeleQuAD \cite{holm2021bidirectional}, a telecom domain QA dataset derived from 3GPP Release 15 documents \cite{3gpp_release_15}. Our experimental setup is shown in Figure \ref{fig:expt-setup}. The input to our pipeline is the QA dataset, which has contexts (from the 3GPP documents) along with associated questions and (ground truth) answers. These questions have been prepared by Subject Matter Experts (SMEs). The training and test data considered comprises of 5,167 and 715 QA, respectively, derived from 452 contexts (sections) from 14 3GPP documents. Appendix \ref{app:sampleQuestions} shows a sample set of QAs along with the contexts. 

 

\subsection{Retriever Models}
\label{ssec:retriever}
The RAG pipeline is comprised of a retriever followed by generator module. The retriever module is comprised of the following steps. Data from reference documents are chunked. An encoder-based language model computes embeddings for query and sentences from the reference documents. For every question embedding, retriever outputs top-$k$ most similar sentence/context embeddings. Cosine similarity is used for selecting top-$k$ sentences/contexts.

We evaluate multiple models in our experiments. From the BAAI family of embedding models, we consider \textit{bge-large-en} \cite{bge_embedding} and \textit{llm-embedder} \cite{llm_embedder}, both with an embedding dimension of $1024$. 
These models have been trained on publicly available datasets; hence, the embeddings may not be optimal for telecom domain. To address this, we also evaluate pre-trained and fine-tuned variants of these models using telecom data. We use sentences from the corpus of technical documents from telecom domain to pre-train \cite{li2020sentence} the base model; we refer to this process as PT in subsequent sections. For fine-tuning \cite{mosbach2020interplay}, we prepare triplets of the form $<q,p,n>$ where $q$ corresponds to the user query, $p$ represents the correct (positive) answer and $n$ is a list of incorrect (negative) answers. The base model is fine-tuned using these triplets; this process is referred to as FT in subsequent experiments. It may be noted here that the fine-tuning may be performed independently on the base model (without pre-training) or post pre-training; in the latter case, we refer to it as PT-FT to  report results. 

\subsection{Generator}
\label{ssec:generator}
The output of the retriever forms the input context to the generator. For evaluation, we only use $k=1$ retrieved context, to study the behaviour of the generator when presented with correct and wrong contexts.
Once the relevant context has been retrieved for a question from the retriever module, the query and context are passed to the LLM for generating the response, indicated as ``RAG Response'' in Figure \ref{fig:expt-setup}. We have considered Mistral-7b \cite{jiang2023mistral} and GPT3.5 as the LLMs for our experiments. We also report results on pre-trained (PT) and instruction fine-tuned (PT-IFT) variants of Mistral-7b using \textit{mistral-finetune} \cite{mistralfinetune}.

\section{RAG Evaluation}
\label{sec:rag_metrics}
We focus on the following metrics from the RAGAS framework \cite{es2023ragas}. Higher value is better for all of them.
\begin{itemize}
    \item 
Faithfulness ($FaiFul$): Checks if the (generated) statements from RAG response are present in the retrieved context
through verdicts; the ratio of valid verdicts to total number of statements in the context is the answer's faithfulness. 
\item Answer Relevance ($AnsRel$): The average cosine similarity of user's question with generated questions, using the RAG response the reference, is the answer relevance. 
\item Context Relevance ($ConRel$): The ratio of the number of statements considered relevant to the question given the context to the total number of statements in the context is the context relevance. 
\item Answer Similarity ($AnsSim$): The similarity between the embedding of RAG response and the embedding of ground truth answer. 
\item Factual Correctness ($FacCor$): This is the F1-Score of statements in RAG response classified as True Positive, False Positive and False Negative by the RAGAS LLM.  
\item Answer Correctness ($AnsCor$): Determines correctness of the generated answer w.r.t. ground truth (as a weighted sum of factual correctness and answer similarity). 
\end{itemize}
 
The RAGAS library has been a black box; hence, interpretability of the scores is difficult as the scores conflicted with human scores by SMEs. To address this, we store the intermediate outputs and verdicts. For details of computation of these metrics, readers can refer to Appendix \ref{app:ragas} and sample output for representative questions in Appendix \ref{app:sampleQuestions}.

We conduct the following experiments to analyse RAG outputs using RAGAS metrics: (i) Compute RAGAS metric on RAG output, (ii) Domain adapt BAAI family of models for retriever in RAG using PT and FT and assess impact on RAGAS metrics, and (iii) Instruction fine tune the LLM for RAG with RAGAS evaluation metrics.

\section{Results and Discussion}
\label{sec:results}

The retriever accuracies for various models {for a range} of $k$ are reported in Table \ref{tab:retr_perf}. The lower accuracy with PT alone is expected and has been discussed in the literature \cite{li2020sentence}. However, these improve significantly ($p<0.05$ for a two-tailed t-test) on FT. We also evaluate with GPT 3.5 (ada-002) embeddings - however, security and privacy concerns limit domain adaptation options for GPT embeddings.
\vspace{-3mm}
\begin{table}[ht]
\centering
\begin{tabular}{|c|r|r|r|}
\hline
Model        & \multicolumn{1}{c|}{k=1} & \multicolumn{1}{c|}{k=3} & \multicolumn{1}{c|}{k=5} \\ \hline \hline
BGE-LARGE    & 69.09                     & 81.64                     & 84.81                     \\ \hline
BGE-PT       & 69.48                     & 79.92                     & 83.36                     \\ \hline
BGE-FT       & 70.67                     & 84.68                     & 88.90                     \\ \hline
BGE-FT-PT    & \textbf{72.66}                     & \textbf{86.79}                     & \textbf{91.02}                     \\ \hline
LLM-EMBEDDER & 68.03                     & 80.71                     & 84.54                     \\ \hline
LLM-PT       & 64.60                     & 75.30                     & 79.66                     \\ \hline
LLM-FT       & 68.96                     & 84.15                     & 88.51                     \\ \hline
LLM-PT-FT    & 70.94                     & 84.54                     & 87.98                     \\ \hline
\end{tabular}

\caption{Retriever performance(\%) using various embedding models for various values of $k$.}
\label{tab:retr_perf}
\end{table}
\vspace{-2mm}



\begin{table*}
\centering
\scalebox{0.75}{
\begin{tabular}{|c|c|c|c|c|c|c|c|c|c|c|}
\hline
\textbf{RAG LLM} & \textbf{RAGAS LLM} & \textbf{Embedding} & \textbf{Retr. Corr.} & \textbf{$FaiFul$}              & \textbf{$AnsRel$}            & \textbf{$ConRel$} & \textbf{$AnsSim$} & \textbf{$AnsCor$} & \textbf{$FacCor$}                      & \textbf{Questions} \\ \hline
Mistral          & Mistral            & BGE BASE           & Yes                  & 0.91(0.19)                         & 0.78(0.09)                      & 0.29(0.28)            & 0.73(0.1)                & 0.76(0.22)          & 0.77(0.29)                                  & 502                \\ \cline{4-11}
                 &                    &                    & No                   & 0.71(0.35)                         & 0.74(0.11)                      & 0.16(0.26)            & 0.61(0.09)               & 0.33(0.27)          & 0.24(0.34)                                  & 213                \\ \hline
Mistral          & Mistral            & BGE FT PT          & Yes                  & 0.91(0.19)                         & 0.55(0.12)                      & 0.28(0.28)            & 0.57(0.15)               & 0.71(0.23)          & 0.76(0.3)                                   & 526                \\ \cline{4-11}
                 &                    &                    & No                   & 0.78(0.31)                         & 0.5(0.13)                       & 0.19(0.27)            & 0.41(0.18)               & 0.3(0.29)           & 0.26(0.35)                                  & 189                \\ \hline
Mistral          & Mistral-IFT        & BGE FT PT          & Yes                  & 0.89(0.26)                         & \multicolumn{1}{l|}{0.36(0.12)} & 0.29(0.28)            & 0.84(0.22)               & 0.82(0.27)          & \textbf{0.82(0.31)} & 526                \\ \cline{4-11}
                 &                    &                    & No                   & 0.68(0.4)                          & \multicolumn{1}{l|}{0.36(0.11)} & 0.19(0.28)            & 0.57(0.26)               & 0.43(0.34)          & 0.38(0.39)          & 189                \\ \hline
Mistral          & Mistral            & LLM                & Yes                  & 0.91(0.18) & \textbf{0.9(0.04)}              & \textbf{0.3(0.29)}    & \textbf{0.88(0.05)}      & 0.79(0.22)          & 0.77(0.29)          & 493                \\ \cline{4-11}
                 &                    &                    & No                   & 0.71(0.35) & 0.88(0.04)                      & 0.15(0.25)            & 0.82(0.04)               & 0.37(0.25)          & 0.22(0.33)          & 222                \\ \hline
Mistral          & Mistral            & LLM PT FT          & Yes                  & 0.91(0.19) & 0.7(0.09)                       & 0.29(0.28)            & 0.68(0.11)               & 0.75(0.23)          & 0.77(0.29)          & 515                \\ \cline{4-11}
                 &                    &                    & No                   & 0.77(0.31) & 0.65(0.09)                      & 0.19(0.28)            & 0.53(0.11)               & 0.32(0.28)          & 0.24(0.35)          & 200                \\ \hline
Mistral          & Mistral-IFT        & LLM PT FT          & Yes                  & 0.88(0.26) & 0.48(0.09)                      & 0.29(0.28)            & 0.88(0.17)               & \textbf{0.83(0.27)} & 0.81(0.31)          & 515                \\ \cline{4-11}
                 &                    &                    & No                   & 0.64(0.42) & 0.48(0.09)                      & 0.19(0.28)            & 0.64(0.21)               & 0.43(0.33)          & 0.36(039)          & 200                \\ \hline
GPT3.5           & GPT 3.5            & BGE FT PT          & Yes                  & \textbf{0.94(0.17)}                & 0.88(0.05)                      & 0.18(0.2)             & 0.87(0.05)               & 0.8(0.21)           & 0.78(0.26)                                  & 526                \\ \cline{4-11}
                 &                    &                    & No                   & 0.68(0.42)                         & 0.83(0.08)                      & 0.13(0.22)            & 0.8(0.06)                & 0.39(0.3)           & 0.26(0.37)                                  & 189                \\ \hline
\end{tabular}
}
\caption{RAGAS Metrics for our dataset with $k=1$ retrieved contexts being passed. The column `Retr. Corr.' indicates if the retrieved context is correct or not. Numbers are mean (s.d.). BGE BASE is the publicly available emebdding for \textit{bge-large-en}, BGE PT FT is the pre-trained finetuned model for the same. Similar notation is followed for \textit{llm-embedder}. Mistral-IFT is the instruction finetuned mistral model. GPT3.5 is the ada-002 embeddings. Values in bold indicate best values of metrics obtained.\label{tab:ragas_metrics_v2}}
\end{table*}

\begin{table*}[ht]
\centering
\scalebox{0.8}{%

\begin{tabular}{|cllllllllll|}
\hline
\multicolumn{11}{|c|}{\textbf{BGE-LARGE}}                                                                                                                                                                                                                                                                                                                                                                                                                                                                                                                                                                                                                                                                                                     \\ \hline
\multicolumn{2}{|c|}{}                                    & \multicolumn{1}{l|}{Retriever} & \multicolumn{2}{c|}{Base}                                                                                                                                       & \multicolumn{2}{c|}{FT}                                                                                                                                         & \multicolumn{2}{c|}{PT}                                                                                                                                         & \multicolumn{2}{c|}{PT-FT}                                                                                                                 \\ \cline{3-11} 
\multicolumn{2}{|c|}{\multirow{-2}{*}{LLM}}               & \multicolumn{1}{l|}{Metric}    & \multicolumn{1}{c|}{$FaiFul$}                                                        & \multicolumn{1}{c|}{$FacCor$}                                                        & \multicolumn{1}{c|}{$FaiFul$}                                                        & \multicolumn{1}{c|}{$FacCor$}                                                        & \multicolumn{1}{c|}{$FaiFul$}                                                        & \multicolumn{1}{c|}{$FacCor$}                                                        & \multicolumn{1}{c|}{$FaiFul$}                                                        & \multicolumn{1}{c|}{$FacCor$}                                   \\ \hline
\multicolumn{2}{|c|}{}                                    & \multicolumn{1}{l|}{Correct}   & \multicolumn{1}{l|}{\cellcolor[HTML]{B5E6A2}\textbf{0.91(0.19)}}               & \multicolumn{1}{l|}{\cellcolor[HTML]{B5E6A2}\textbf{0.77(0.29)}}               & \multicolumn{1}{l|}{\cellcolor[HTML]{FFFFFF}0.91(0.18)}                        & \multicolumn{1}{l|}{\cellcolor[HTML]{FFFFFF}0.76(0.29)}                        & \multicolumn{1}{l|}{\cellcolor[HTML]{FFFFFF}0.90(0.19)}                         & \multicolumn{1}{l|}{\cellcolor[HTML]{FFFFFF}0.76(0.30)}                         & \multicolumn{1}{l|}{\cellcolor[HTML]{FFFFFF}0.91(0.19)}                        & \cellcolor[HTML]{FFFFFF}0.76(0.30)                         \\ \cline{3-11} 
\multicolumn{2}{|c|}{\multirow{-2}{*}{Mistral-7b}}        & \multicolumn{1}{l|}{Wrong}     & \multicolumn{1}{l|}{\cellcolor[HTML]{B5E6A2}\textbf{0.71(0.35)}}               & \multicolumn{1}{l|}{\cellcolor[HTML]{B5E6A2}\textbf{0.24(0.34)}}               & \multicolumn{1}{l|}{\cellcolor[HTML]{FFFFFF}0.76(0.31)}                        & \multicolumn{1}{l|}{\cellcolor[HTML]{FFFFFF}0.25(0.34)}                        & \multicolumn{1}{l|}{\cellcolor[HTML]{FFFFFF}{\color[HTML]{0066FF} 0.78(0.31)}} & \multicolumn{1}{l|}{\cellcolor[HTML]{FFFFFF}0.28(0.36)}                        & \multicolumn{1}{l|}{\cellcolor[HTML]{FFFFFF}{\color[HTML]{0066FF} 0.78(0.31)}} & \cellcolor[HTML]{FFFFFF}0.26(0.35)                        \\ \hline
\multicolumn{2}{|c|}{}                                    & \multicolumn{1}{l|}{Correct}   & \multicolumn{1}{l|}{\cellcolor[HTML]{FFFFFF}0.89(0.26)}                        & \multicolumn{1}{l|}{\cellcolor[HTML]{FFFFFF}{\color[HTML]{0066FF} 0.82(0.30)}}  & \multicolumn{1}{l|}{\cellcolor[HTML]{FFFFFF}0.88(0.26)}                        & \multicolumn{1}{l|}{\cellcolor[HTML]{FFFFFF}{\color[HTML]{0066FF} 0.82(0.31)}} & \multicolumn{1}{l|}{\cellcolor[HTML]{FFFFFF}0.89(0.26)}                        & \multicolumn{1}{l|}{\cellcolor[HTML]{FFFFFF}{\color[HTML]{0066FF} 0.80(0.32)}}  & \multicolumn{1}{l|}{\cellcolor[HTML]{FFFFFF}0.89(0.25)}                        & \cellcolor[HTML]{FFFFFF}{\color[HTML]{0066FF} 0.82(0.30)}  \\ \cline{3-11} 
\multicolumn{2}{|c|}{\multirow{-2}{*}{Mistral-7b-IFT}}    & \multicolumn{1}{l|}{Wrong}     & \multicolumn{1}{l|}{\cellcolor[HTML]{FFFFFF}{\color[HTML]{FF0000} 0.58(0.44)}} & \multicolumn{1}{l|}{\cellcolor[HTML]{FFFFFF}{\color[HTML]{0066FF} 0.36(0.39)}} & \multicolumn{1}{l|}{\cellcolor[HTML]{FFFFFF}{\color[HTML]{FF0000} 0.64(0.43)}} & \multicolumn{1}{l|}{\cellcolor[HTML]{FFFFFF}{\color[HTML]{0066FF} 0.37(0.39)}} & \multicolumn{1}{l|}{\cellcolor[HTML]{FFFFFF}{\color[HTML]{FF0000} 0.57(0.44)}} & \multicolumn{1}{l|}{\cellcolor[HTML]{FFFFFF}{\color[HTML]{0066FF} 0.36(0.39)}} & \multicolumn{1}{l|}{\cellcolor[HTML]{FFFFFF}0.66(0.41)}                        & \cellcolor[HTML]{FFFFFF}{\color[HTML]{0066FF} 0.38(0.39)} \\ \hline
\multicolumn{2}{|c|}{}                                    & \multicolumn{1}{l|}{Correct}   & \multicolumn{1}{l|}{\cellcolor[HTML]{FFFFFF}0.88(0.26)}                        & \multicolumn{1}{l|}{\cellcolor[HTML]{FFFFFF}{\color[HTML]{0066FF} 0.80(0.32)}}  & \multicolumn{1}{l|}{\cellcolor[HTML]{FFFFFF}0.88(0.26)}                        & \multicolumn{1}{l|}{\cellcolor[HTML]{FFFFFF}0.80(0.32)}                         & \multicolumn{1}{l|}{\cellcolor[HTML]{FFFFFF}0.89(0.25)}                        & \multicolumn{1}{l|}{\cellcolor[HTML]{FFFFFF}{\color[HTML]{0066FF} 0.79(0.33)}} & \multicolumn{1}{l|}{\cellcolor[HTML]{FFFFFF}0.88(0.26)}                        & \cellcolor[HTML]{FFFFFF}{\color[HTML]{0066FF} 0.80(0.32)}  \\ \cline{3-11} 
\multicolumn{2}{|c|}{\multirow{-2}{*}{Mistral-7b-PT-IFT}} & \multicolumn{1}{l|}{Wrong}     & \multicolumn{1}{l|}{\cellcolor[HTML]{FFFFFF}{\color[HTML]{FF0000} 0.59(0.44)}} & \multicolumn{1}{l|}{\cellcolor[HTML]{FFFFFF}{\color[HTML]{0066FF} 0.36(0.38)}} & \multicolumn{1}{l|}{\cellcolor[HTML]{FFFFFF}{\color[HTML]{FF0000} 0.65(0.42)}} & \multicolumn{1}{l|}{\cellcolor[HTML]{FFFFFF}{\color[HTML]{0066FF} 0.35(0.39)}} & \multicolumn{1}{l|}{\cellcolor[HTML]{FFFFFF}{\color[HTML]{FF0000} 0.61(0.44)}} & \multicolumn{1}{l|}{\cellcolor[HTML]{FFFFFF}{\color[HTML]{0066FF} 0.35(0.39)}} & \multicolumn{1}{l|}{\cellcolor[HTML]{FFFFFF}0.65(0.43)}                        & \cellcolor[HTML]{FFFFFF}{\color[HTML]{0066FF} 0.36(0.39)} \\ \hline
\multicolumn{11}{|c|}{\textbf{LLM-EMBEDDER}}                                                                                                                                                                                                                                                                                                                                                                                                                                                                                                                                                                                                                                                                                                  \\ \hline
\multicolumn{2}{|c|}{}                                    & \multicolumn{1}{l|}{Retriever} & \multicolumn{2}{c|}{Base}                                                                                                                                       & \multicolumn{2}{c|}{FT}                                                                                                                                         & \multicolumn{2}{c|}{PT}                                                                                                                                         & \multicolumn{2}{c|}{PT-FT}                                                                                                                 \\ \cline{3-11} 
\multicolumn{2}{|c|}{\multirow{-2}{*}{LLM}}               & \multicolumn{1}{l|}{Metric}    & \multicolumn{1}{c|}{$FaiFul$}                                                        & \multicolumn{1}{c|}{$FacCor$}                                                        & \multicolumn{1}{c|}{$FaiFul$}                                                        & \multicolumn{1}{c|}{$FacCor$}                                                        & \multicolumn{1}{c|}{$FaiFul$}                                                        & \multicolumn{1}{c|}{$FacCor$}                                                        & \multicolumn{1}{c|}{$FaiFul$}                                                        & \multicolumn{1}{c|}{$FacCor$}                                   \\ \hline
\multicolumn{2}{|c|}{}                                    & \multicolumn{1}{l|}{Correct}   & \multicolumn{1}{l|}{\cellcolor[HTML]{B5E6A2}\textbf{0.91(0.18)}}               & \multicolumn{1}{l|}{\cellcolor[HTML]{B5E6A2}\textbf{0.77(0.29)}}               & \multicolumn{1}{l|}{\cellcolor[HTML]{FFFFFF}0.92(0.18)}                        & \multicolumn{1}{l|}{\cellcolor[HTML]{FFFFFF}0.77(0.29)}                        & \multicolumn{1}{l|}{\cellcolor[HTML]{FFFFFF}0.91(0.2)}                         & \multicolumn{1}{l|}{\cellcolor[HTML]{FFFFFF}0.76(0.30)}                         & \multicolumn{1}{l|}{\cellcolor[HTML]{FFFFFF}0.91(0.19)}                        & \cellcolor[HTML]{FFFFFF}0.77(0.29)                        \\ \cline{3-11} 
\multicolumn{2}{|c|}{\multirow{-2}{*}{Mistral-7b}}        & \multicolumn{1}{l|}{Wrong}     & \multicolumn{1}{l|}{\cellcolor[HTML]{B5E6A2}\textbf{0.71(0.35)}}               & \multicolumn{1}{l|}{\cellcolor[HTML]{B5E6A2}\textbf{0.22(0.33)}}               & \multicolumn{1}{l|}{\cellcolor[HTML]{FFFFFF}0.72(0.34)}                        & \multicolumn{1}{l|}{\cellcolor[HTML]{FFFFFF}0.22(0.33)}                        & \multicolumn{1}{l|}{\cellcolor[HTML]{FFFFFF}0.73(0.33)}                        & \multicolumn{1}{l|}{\cellcolor[HTML]{FFFFFF}0.19(0.32)}                        & \multicolumn{1}{l|}{\cellcolor[HTML]{FFFFFF}{\color[HTML]{0066FF} 0.77(0.31)}} & \cellcolor[HTML]{FFFFFF}0.24(0.34)                        \\ \hline
\multicolumn{2}{|c|}{}                                    & \multicolumn{1}{l|}{Correct}   & \multicolumn{1}{l|}{\cellcolor[HTML]{FFFFFF}0.88(0.26)}                        & \multicolumn{1}{l|}{\cellcolor[HTML]{FFFFFF}{\color[HTML]{0066FF} 0.81(0.31)}} & \multicolumn{1}{l|}{\cellcolor[HTML]{FFFFFF}0.89(0.25)}                        & \multicolumn{1}{l|}{\cellcolor[HTML]{FFFFFF}{\color[HTML]{0066FF} 0.81(0.31)}} & \multicolumn{1}{l|}{\cellcolor[HTML]{FFFFFF}0.88(0.26)}                        & \multicolumn{1}{l|}{\cellcolor[HTML]{FFFFFF}{\color[HTML]{0066FF} 0.81(0.31)}} & \multicolumn{1}{l|}{\cellcolor[HTML]{FFFFFF}0.89(0.26)}                        & \cellcolor[HTML]{FFFFFF}{\color[HTML]{0066FF} 0.81(0.31)} \\ \cline{3-11} 
\multicolumn{2}{|c|}{\multirow{-2}{*}{Mistral-7b-IFT}}    & \multicolumn{1}{l|}{Wrong}     & \multicolumn{1}{l|}{\cellcolor[HTML]{FFFFFF}{\color[HTML]{FF0000} 0.55(0.45)}} & \multicolumn{1}{l|}{\cellcolor[HTML]{FFFFFF}{\color[HTML]{0066FF} 0.32(0.38)}} & \multicolumn{1}{l|}{\cellcolor[HTML]{FFFFFF}{\color[HTML]{FF0000} 0.57(0.45)}} & \multicolumn{1}{l|}{\cellcolor[HTML]{FFFFFF}{\color[HTML]{0066FF} 0.36(0.39)}} & \multicolumn{1}{l|}{\cellcolor[HTML]{FFFFFF}{\color[HTML]{FF0000} 0.56(0.44)}} & \multicolumn{1}{l|}{\cellcolor[HTML]{FFFFFF}{\color[HTML]{0066FF} 0.32(0.38)}} & \multicolumn{1}{l|}{\cellcolor[HTML]{FFFFFF}{\color[HTML]{FF0000} 0.63(0.44)}} & \cellcolor[HTML]{FFFFFF}{\color[HTML]{0066FF} 0.36(0.39)} \\ \hline
\multicolumn{2}{|c|}{}                                    & \multicolumn{1}{l|}{Correct}   & \multicolumn{1}{l|}{\cellcolor[HTML]{FFFFFF}0.88(0.25)}                        & \multicolumn{1}{l|}{\cellcolor[HTML]{FFFFFF}{\color[HTML]{0066FF} 0.80(0.33)}}  & \multicolumn{1}{l|}{\cellcolor[HTML]{FFFFFF}0.88(0.26)}                        & \multicolumn{1}{l|}{\cellcolor[HTML]{FFFFFF}{\color[HTML]{0066FF} 0.80(0.32)}}  & \multicolumn{1}{l|}{\cellcolor[HTML]{FFFFFF}0.87(0.27)}                        & \multicolumn{1}{l|}{\cellcolor[HTML]{FFFFFF}0.79(0.33)}                        & \multicolumn{1}{l|}{\cellcolor[HTML]{FFFFFF}0.88(0.25)}                        & \cellcolor[HTML]{FFFFFF}0.80(0.33)                         \\ \cline{3-11} 
\multicolumn{2}{|c|}{\multirow{-2}{*}{Mistral-7b-PT-IFT}} & \multicolumn{1}{l|}{Wrong}     & \multicolumn{1}{l|}{\cellcolor[HTML]{FFFFFF}{\color[HTML]{FF0000} 0.55(0.45)}} & \multicolumn{1}{l|}{\cellcolor[HTML]{FFFFFF}{\color[HTML]{0066FF} 0.33(0.38)}} & \multicolumn{1}{l|}{\cellcolor[HTML]{FFFFFF}{\color[HTML]{FF0000} 0.62(0.44)}} & \multicolumn{1}{l|}{\cellcolor[HTML]{FFFFFF}{\color[HTML]{0066FF} 0.34(0.38)}} & \multicolumn{1}{l|}{\cellcolor[HTML]{FFFFFF}{\color[HTML]{FF0000} 0.56(0.44)}} & \multicolumn{1}{l|}{\cellcolor[HTML]{FFFFFF}{\color[HTML]{0066FF} 0.32(0.38)}} & \multicolumn{1}{l|}{\cellcolor[HTML]{FFFFFF}{\color[HTML]{FF0000} 0.64(0.43)}} & \cellcolor[HTML]{FFFFFF}{\color[HTML]{0066FF} 0.35(0.39)} \\ \hline
\end{tabular}
}
\caption{Results with Instruction Fine-tuned LLMs (Mistral-7b), cells highlighted in green indicate baseline results (with base version of embedding model and LLM). Numbers in blue and red indicate results that are statistically significant w.r.t. baseline results. Blue and red indicate statistically significant ($p <0.05 $) increase and decrease w.r.t. baseline results, respectively. Numbers are mean (s.d.).}
\label{tab:ragas_metrics_ift}
\end{table*}

The results of the RAG evaluation are shown in Table \ref{tab:ragas_metrics_v2}. 
We include scores for the sub-components of $AnsCor$ i.e., Answer Similarity ($AnsSim$) and Factual Correctness ($FacCor$); $AnsCor$ is their weighted average with weights $0.25$ and $0.75$ respectively. 

We note that the mean of the considered metrics for \textit{Retrieval correct=`Yes'} is greater than or equal to that for \textit{Retrieval correct=`No'} (validated by one-sided t-test, $p<0.05$ for statistical significance). Next, we observe that for $FaiFul$ and $AnsCor$, the results for PT and FT are similar to that of the base model ($p>0.05$). 
The other metrics are not truly comparable (discussed in detail in Section \ref{subsec:metricDiscussion}). 
We observe that the metrics values reported using open source LLM and GPT3.5 are comparable. Instruction Fine Tuning of the generator improves the relevant metrics.


\subsection{ Discussion on Metrics}\label{subsec:metricDiscussion}

We discuss our findings about the four RAGAS metrics.

\begin{itemize}
    \item \textbf{Faithfulness ($FaiFul$)} - intends to provide reliability scores with respect to human evaluation. Simple statements might be paraphrased into multiple sentences, while complex statements may not be fully broken down. These factors can introduce variation in the faithfulness score. Despite these challenges, we found that the $FaiFul$ metric is generally concordant to manual evaluation.
    \item \textbf{Context Relevance ($ConRel$)} - is mainly indicative and dependent on the context length. Typically, chunks such as sections form the retrieved context in a RAG pipeline; hence, context can vary in length, which affects the denominator component of $ConRel$. Typical metrics are either ``higher is better" (e.g., accuracy) or ``lower is better" (e.g., mean squared error). Also, all statements are assigned equal weight, regardless of the length or quality of the sentence. Therefore, we infer $ConRel$ cannot be appropriately clubbed into either of these types and the final metric is hard to interpret or even have an intuition about.
    \item \textbf{Answer Relevance ($AnsRel$)} - The generated questions from the LLM in this metric may not be the best way to measure answer relevance. We have observed some cases where the generated questions are either trivial paraphrasing or incorrect. However, the major problem with this metric is that it tries to use cosine similarity as an absolute component of this metric. This makes the metric dependent on the choice of LLM. In addition, various studies on cosine similarity have pointed out that it may not be indicative of similarity \cite{steck2024cosine}, known to give artificially low values \cite{fourmetrics}, is difficult to provide thresholds on \cite{zhu2010top} and is subject to the isotropy of embeddings \cite{timkey2021all}. For many embeddings, similarity can be very high even between random words/sentences \cite{ethayarajh2019contextual}. All this points to the fact that using cosine similarity as a metric, like is done for $AnsRel$, is not very interpretable.
    \item \textbf{Answer Correctness ($AnsCor$)} - The FC component of this score is dependent on the LLM correctly identifying the True Positives (TP), False Positives (FP), and False Negatives (FN). Our analysis shows that this process can sometimes result in incorrect mapping of sentences to these groups. Occasionally, irrelevant statements might be included, or relevant sentences might not be classified into any of the groups. Additionally, the semantic correctness component of this score being a cosine similarity is subject to the concerns raised on the $AnsRel$ metric. Despite this, our manual analysis shows that the answer correctness score is relatively well aligned with a human expert evaluation.
\end{itemize}

In summary, our results indicate that of these metrics, $FaiFul$ and $AnsCor$ are perhaps best aligned with human expert judgment; scores for $AnsSim$, $AnsRel$ and $ConRel$ are subject to inherent variations and are relatively unreliable for interpretation. Hence, we report results in more detail for only $FaiFul$ and $FacCor$ (i.e., $AnsCor = FacCor$ with weight for $AnsSim$ set to $0$) in Table \ref{tab:ragas_metrics_ift}. For correct retrieval, both $FaiFul$ and $FacCor$ are as good or better ($p<0.05$) with domain adaptation as expected. For wrong retrieval, an improvement in $FaiFul$ should necessarily lead to a reduction in $FacCor$ and vice-versa. We observe that the generator LLM is answering questions from it's enriched domain adapted knowledge, leading to a lower $FaiFul$ and higher $FacCor$. Although not desirable from the expected generator response, the metrics correctly captures this. 

 Further, the RAG responses are evaluated for correctness by  SMEs, all responses from each of Mistral-7b, Mistral-7b-IFT and Mistral-7b-PT-IFT. 
 Considering this evaluation as ground truth for correctness, we evaluate the probability of the answer being correct based on the RAGAS metrics.
 
  For each of $FaiFul$ and $FacCor$ metrics, we compute the probability of correct generated answer considering both the metrics $(m1,m2) \in \{FaiFul,FacCor\}$ being above a certain threshold, using Equations (\ref{eqn:condProbs3}) - (\ref{eqn:condProbs4}). 



\begin{equation}
\begin{aligned}
&P(c|m_1>\theta_{11};m_2>\theta_{12}) \\
&\quad = \frac{P(m_1>\theta_{11};m_2>\theta_{12}|c)P(c)}{P(m_1>\theta_{11};m_2>\theta_{12})}
\end{aligned}
\label{eqn:condProbs3}
\end{equation}

\begin{equation}
\begin{aligned}
&P(w|m_1<\theta_{21};m_2<\theta_{22}) \\
&\quad = \frac{P(m_1<\theta_{21};m_2<\theta_{22}|w)P(w)}{P(m_1<\theta_{21};m_2<\theta_{22})}
\end{aligned}
\label{eqn:condProbs4}
\end{equation}
It is possible to use the Bayesian formulae with only one metric considered instead of the joint conditional distribution. 
Table \ref{tab:concordance} shows the results for $\theta_{11}=\theta_{12}=0.7$ and $\theta_{21}=\theta_{22}=0.3$ considering each metric independently and both jointly. We observe better concordance of RAGAS metrics with that of SME evaluation, if both the metrics are considered together. We also observe that domain adaptation of the LLM via IFT and PT-IFT improves the scores significantly ($p<0.05$). However, there is little difference ($p>0.05$) between IFT alone and PT-IFT.
 We conclude that these two metrics are well aligned with human judgement and can be used in an end-to-end pipeline reliably. 
 
 Sample outputs of RAGAS metrics for some questions and the corresponding metrics are shown in Appendix  \ref{app:sampleQuestions}.

\begin{table}[t]
\centering
\resizebox{0.98\columnwidth}{!}{%
\begin{tabular}{|l|c|r|r|r|}
\hline
\textbf{Metric} & \textbf{LLM Model} & \textbf{$FacCor$} & \textbf{$FaiFul$} & \textbf{Joint} \\ \hline
\multirow{3}{*}{$\begin{aligned}P(c|m_1>\theta_{11} \\ ;m_2>\theta_{12})\end{aligned}$} & Mistral 7B        & 0.87 & 0.74 & 0.87 \\ \cline{2-5}
   & Mistral 7B IFT    & 0.96 & 0.76 & 0.97 \\ \cline{2-5}
   & Mistral 7B PT IFT & 0.96 & 0.79 & 0.97 \\ \hline
\multirow{3}{*}{$\begin{aligned}P(w|m_1<\theta_{21} \\ ;m_2<\theta_{22})\end{aligned}$} & Mistral 7B        & 0.72 & 0.71 & 0.84 \\ \cline{2-5}
   & Mistral 7B IFT    & 0.79 & 0.70 & 0.86 \\ \cline{2-5}
   & Mistral 7B PT IFT & 0.75 & 0.75 & 0.89 \\ \hline
\end{tabular}%
}
\caption{Concordance of selected metrics with expert evaluation of correctness (embedder is \textit{bge-large})\label{tab:concordance}. The shown threshold is the same for both metrics i.e. $\theta_{11}=\theta_{12}=0.7$ and $\theta_{21}=\theta_{22}=0.3$}.
\end{table}



\section{Conclusions and Future Work}

In this work, we enhance the current version of the RAGAS package for evaluation of RAG based QA - this helps investigate the scores by analysing the intermediate outputs. We focus our study using telecom domain QA. We critique $AnsSim$ component of $AnsCor$, $ConRel$ and $AnsRel$ metrics for their lack of suitability as a reliable metric in an end-to-end RAG pipeline. A detailed analysis by SMEs of RAG output establishes that two of the metrics $FacCor$ and $FaiFul$ are suitable for evaluation purposes in RAG pipeline. 
We demonstrate that domain adaptation of RAG LLM improves the concordance of the two metrics with SME evaluations. Whilst our studies have been limited to telecom domain, some of our concerns especially around the use of cosine similarity would extend to other domains too.

Our code repository presents the intermediate output of the RAGAS metrics, and possibilities for improvements in RAG evaluation across  domains. A detailed study of other libraries dependent on RAGAS like ARES \cite{saad2023ares} can also be considered in future.

\label{sec:conclusion}
\newpage
\bibliography{rag_fm_wild_ragas}
\bibliographystyle{icml2024}

\newpage
\appendix
\onecolumn
\section*{Appendices}


\section{Computation of RAGAS Metrics} \label{app:ragas}
\begin{figure}[h]
    \centering
    \includegraphics[width=0.95\linewidth]{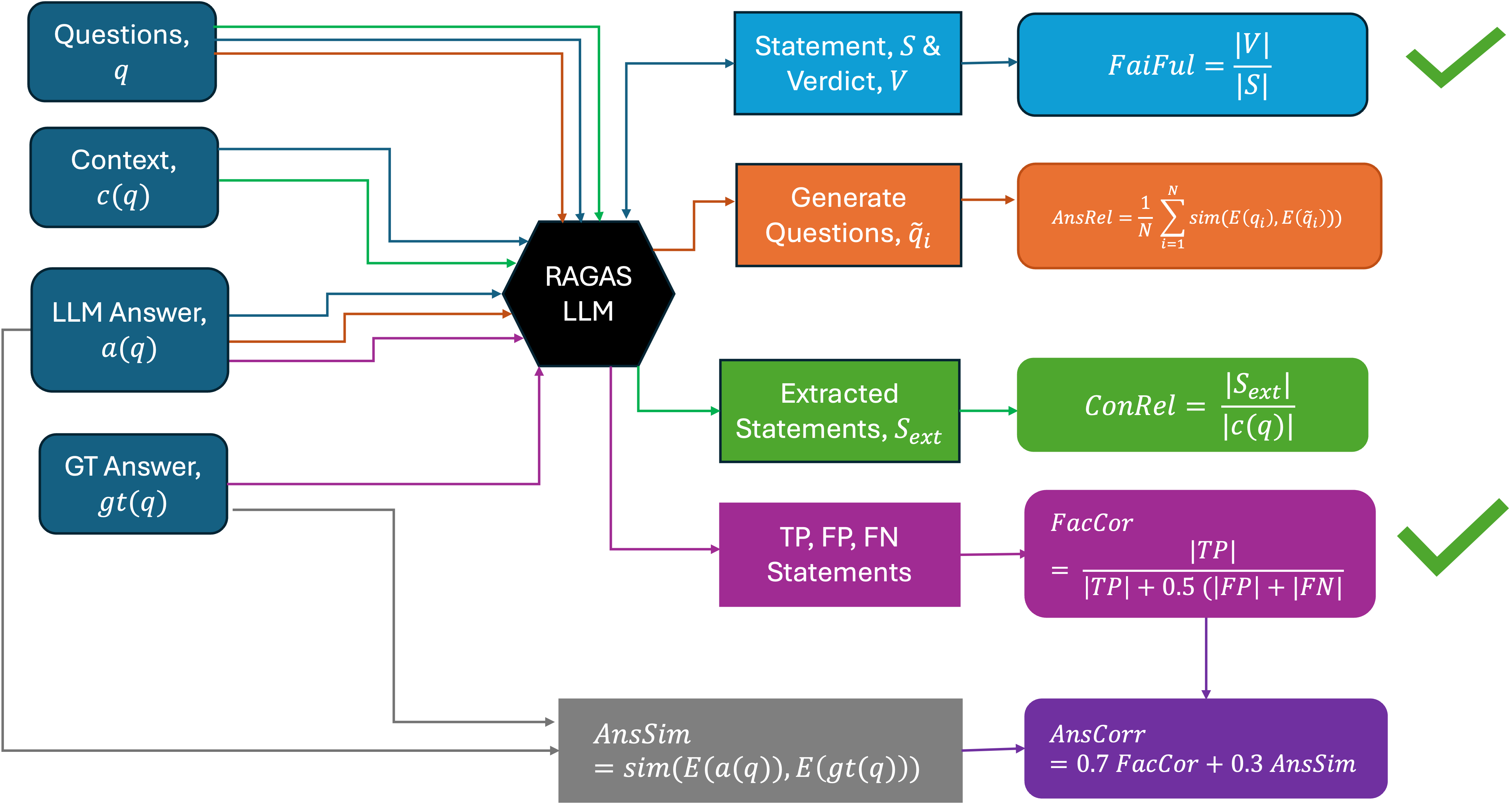}
    \caption{Summary view of RAGAS Metrics and their computation. Green check mark indicates recommended metrics, based on our experiments.}
    \label{fig:ragas_metrics_diagram}
\end{figure}

We refer the reader to \cite{es2023ragas} for details on the metrics defined, but for the sake of completeness, the prompts involved and steps to determine the metrics in our study are provided for ease of reference. A summary is shown in Figure \ref{fig:ragas_metrics_diagram}. The notation used is as follows: given question $q$ and context $c(q)$ retrieved (and possibly re-ranked) from a corpus, the LLM generates answer $a(q)$. The ground truth answer for the question is denoted by $gt(q)$.

\subsection{Faithfulness ($FaiFul$)}
By definition, answer $a(q)$ is faithful to context $c(q)$ \textit{``if claims made in the answer can be inferred from the context''}. This is done using a two-step process. In the first step, the LLM is prompted to create sentences (statements $S(q)$) using the answer $a(q)$ using the following prompt:

\begin{quote}
    Given a question and answer, create one or more statements from each sentence in the given answer.\\
    question: [question] \\
    answer: [answer]
\end{quote}

In the next step, for each statement $s \in S(q)$, the LLM is asked to determine a binary verdict $v(s,c(q))$ using the context as part of the following prompt:

\begin{quote}
    Consider the given context and following statements, then determine whether they are supported by the information present in the context. Provide a brief explanation for each statement before arriving at the verdict (Yes/No). Provide a final verdict for each statement in order at the end in the given format. Do not deviate from the specified format. \\
context: [context] \\
statement: [statement 1] ...\\
statement: [statement n] 
\end{quote}

Once the set of verdicts $V$ are obtained, faithfulness $(FaiFul)$ is computed as 
\begin{gather}
    FaiFul = \frac{|V|}{|S|}
\end{gather}
\vspace{-5mm}
\subsection{Answer Relevance ($AnsRel$)}

Answer $a(q)$ \textit{``is relevant if it directly addresses the question in an appropriate way''.} To determine answer correctness, the LLM is used to generate questions $\Tilde{q}$ from the answer $a(q)$ using the following prompt:
\begin{quote}
    Generate a question for the given answer. \\
    answer: [answer] 
\end{quote}

Following this, the similarity of the $N$ generated questions in $\Tilde{q}$ with the original question $q$ is determined using a similarity function $sim(\cdot)$, that takes the embedding $E(\cdot)$ generated by a suitable model as input, and the average similarity score is reported as the answer relevance ($AnsRel$) using
\begin{gather}
    AnsRel = \frac{1}{N}\sum_{i=1}^N sim(E(q),E(\Tilde{q_i}))
\end{gather}
\vspace{-5mm}
\subsection{Context Relevance ($ConRel$)}
By definition, \textit{``the context $c(q)$ is considered relevant to the extent that it exclusively contains information that is  needed to answer the question.''}. This is accomplished by prompting the LLM to extract relevant sentences $S_{ext}$ from the question $q$ and context $c(q)$ using the following prompt:
\begin{quote}
    Please extract relevant sentences from the provided context that can potentially help answer the following question. If no relevant sentences are found, or if you believe the question cannot be answered from the given context, return the phrase ``Insufficient Information". While extracting candidate sentences you’re not allowed to make any changes to sentences from given context. \\
question:  [question] \\
context: [context]
\end{quote}

Following this, context relevance ($ConRel$) is computed as
\begin{gather}
    ConRel = \frac{|S_{ext}|}{|c(q)|},
\end{gather}
\noindent where $|\cdot|$ represents the number of sentences.
\vspace{-3mm}
\subsection{Answer Similarity ($AnsSim$)}
Answer Similarity ($AnsSim$) is defined as the similarity between the LLM generated response $a(q)$ and the ground truth answer $gt(q)$, and is computed using
\begin{gather}
    AnsSim = sim(E(a(q)),E(gt(q)))
\end{gather}
\vspace{-8mm}
\subsection{Answer Correctness ($AnsCor$)}

To determine answer correctness, $a(q)$ and $gt(q)$ are used to generate the following sets of statements:
\begin{itemize}
    \item TP (True Positive): Facts or statements that are present in both the ground truth and the generated answer.
\item FP (False Positive): Facts or statements that are present in the generated answer but not in the ground truth.
\item FN (False Negative): Facts or statements that are present in the ground truth but not in the generated answer.
\end{itemize}

Using these statements, the Factual Correctness ($FacCor$) score is determined as
\begin{gather}
    FacCor = \frac{|TP|}{|TP| + 0.5 \times (|FP| + |FN|) }
\end{gather}
The prompt used to generate $FacCor$ is as follows: 
\begin{quote}
Extract following from given question and ground truth.
``TP": statements that are present in both the answer and the ground truth,``FP": statements present in the answer but not found in the ground truth,``FN": relevant statements found in the ground truth but omitted in the answer.

question: [question],

answer: [answer],  

ground truth: [ground truth answer],

Extracted statements: {

    ``TP": [statement 1, statement 4, \ldots],
    
    ``FP": [statement 2, \ldots],
    
    ``FN": [statement 3, statement 5, statement 6, \ldots]
}

\end{quote}

The answer correctness ($AnsCor$) is defined as the weighted sum of $FC$ score and $AS$ i.e.,
\begin{gather}
    AnsCor = w_1 \times FacCor + w_2 \times AnsSim
\end{gather}
with default weights as per RAGAS implementation $[w_1, w_2] = [0.75, 0.25]$.

\newpage
\section{Sample Questions and RAGAS Metrics with Supporting Statements}\label{app:sampleQuestions}
\begin{figure}[h]
    \centering
    \includegraphics[width=0.93\linewidth]{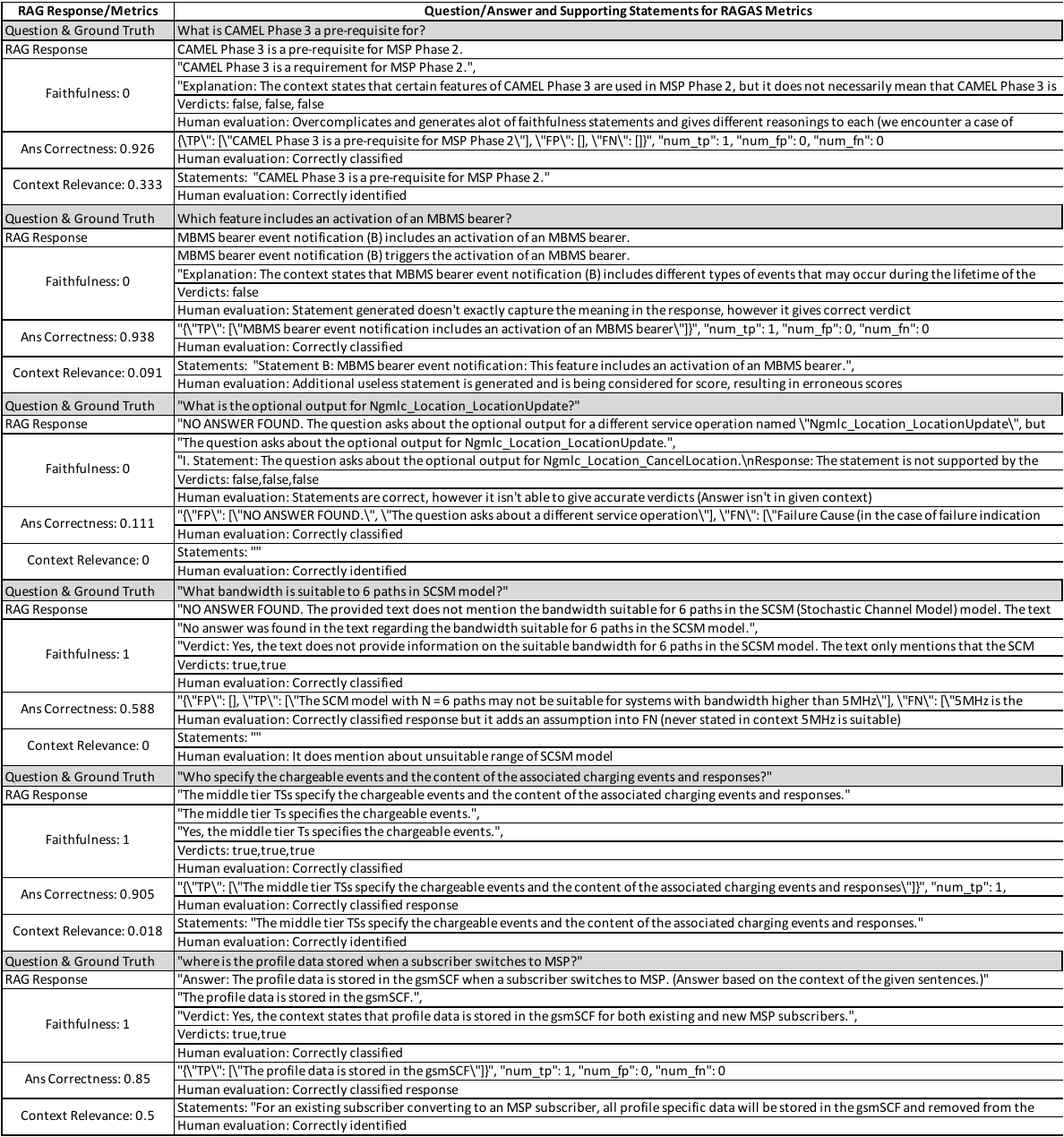}
    \caption{Sample Questions}
    \label{fig:sample_questions}
\end{figure}

Some sample questions and the RAGAS metrics with intermediate outputs are shown in Figure \ref{fig:sample_questions}.



\end{document}